\documentclass{article}
\usepackage[T1]{fontenc}
\usepackage[margin=1in]{geometry}
\usepackage[verbose=silent]{microtype}
\usepackage{hyperref}
\usepackage[numbers]{natbib}
\usepackage{graphicx}
\usepackage{tabularx}
\usepackage{afterpage}
\usepackage{indentfirst}

\newcolumntype{L}{>{\raggedright\arraybackslash}X}

\renewenvironment{abstract}{%
  \small
  \begin{center}%
    {\bfseries\abstractname\vspace{-.5em}\vspace{0pt}}%
  \end{center}%
  \list{}{%
    \setlength{\leftmargin}{1.5cm}%
    \setlength{\rightmargin}{\leftmargin}%
  }%
  \item\relax
}{%
  \endlist
}

\title{In harmony\raisebox{-0.2ex}{\includegraphics[height=1.2em]{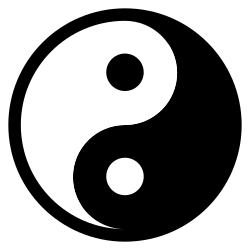}} with gpt-oss}
\author{Borislav Mavrin \\ {\small \texttt{borislav.mavrin@gmail.com}}}
\date{}
\begin{document}
    \maketitle
    \begin{abstract}
      No one has independently reproduced OpenAI's published scores for \texttt{gpt-oss-20b} with tools\footnote{To the best of our knowledge, as of March 2026}, because the original paper discloses neither the tools nor the agent harness.
      We reverse-engineered the model's in-distribution tools: when prompted without tool definitions, \texttt{gpt-oss} still calls tools from its training distribution with high statistical confidence -- a strong prior, not a hallucination.
      We then built a native harmony agent harness\footnote{https://github.com/borislavmavrin/harmonyagent.git} that encodes messages in the model's native format, bypassing the lossy Chat Completions conversion.
      Together, these yield the first independent reproduction of OpenAI's published scores: 60.4\% on SWE Verified HIGH (published 60.7\%), 53.3\% MEDIUM (53.2\%), and 91.7\% on AIME25 with tools (90.4\%).
    \end{abstract}

    \section*{Introduction}

    Many open-source models run on consumer PCs, but only a few can solve coding tasks well: \texttt{Devstral Small 2}, \texttt{GLM-4.7-Flash}, \texttt{Qwen3.5 27b+}, \texttt{gpt-oss-20b}.

    Practitioners naturally pick the largest model size that "fits" on their GPU.
    However, the word "fits" deceives.
    For example, OpenAI claims that "gpt-oss-20b only requires 16GB."\cite{gpt-oss-intro}
    However, even if GPU VRAM is enough to load the model weights, a coding agent requires more.
    Coding tasks require long contexts of 128k+, which demands more VRAM.
    Keeping the KV cache across agent turns rather than recomputing it from scratch requires even more VRAM.
    So 16GB falls short for agentic use.
    These memory demands further constrain which LLM a practitioner can choose for coding agent tasks.
    
    Latency and throughput also matter.
    Parameter count alone does not predict them.
    Architecture matters too -- a dense model like \texttt{Devstral Small 2} requires more compute than an equally sized MoE model, which activates fewer parameters per forward pass.

    Among efficient models that "fit" on a consumer GPU, we then choose by coding capability.
    SWE-bench is among the most meaningful\footnote{There is another pitfall - dataset contamination problem of SWE Verified\cite{swe-contamination} - which we only acknowledge here, but we leave it outside the scope of the paper.} benchmarks because it mirrors real workflows: models must read files, run tests, and edit code through a shell across multiple turns\footnote{Which can easily exceed 100 steps.}.
    
    Given these factors, \texttt{gpt-oss-20b} still offers the best balance of published benchmarks and efficiency among viable local coding models (Table~\ref{tab:models_table}), even though it was released in August 2025.

    However, published SWE Verified scores may not reproduce under a different agent harness.
    SWE-bench does not specify an agent harness -- the scaffolding that connects a model to tools like \texttt{bash}.
    Each provider builds its own harness, most of them proprietary, and the harness alone can shift scores by tens of percentage points.
    \texttt{Devstral Small 2} scores 68\% under its vendor harness but only 56.4\% under the open-source harness on the official SWE leaderboard \cite{mini-swe-agent}.
    For \texttt{gpt-oss-120b}, the gap is even wider: 62.4\% published versus 26\% on the leaderboard\footnotemark[5].

    \begin{table}[!htbp]
        \centering
        \begin{tabularx}{\linewidth}{LLLLL}
          \hline
          \textbf{Model} & \textbf{Published SWE verified score} & \textbf{SWE Leaderboard score}\footnotemark[5] & \textbf{Model size} & \textbf{Model architecture} \\
          \hline
          Devstral Small 2 & 68.0\%\cite{devstralsmall2} & 56.4\% & 24B & Dense \\
          GLM-4.7-Flash & 59.2\%\cite{glm47flash} & N/A & 31B & MoE \\
          Qwen3.5 27b & 72.4\%\cite{qwen3.527B} & N/A & 27B & Dense \\
          gpt-oss-20b & 60.7\%\cite{gptoss} & 26\%\footnotemark[6] & 20B & MoE \\
          \hline
        \end{tabularx}
        \caption{Model size vs SWE Verified scores}
        \label{tab:models_table}
    \end{table}
    \afterpage{%
      \footnotetext[5]{\url{https://www.swebench.com}}%
      \footnotetext[6]{Leaderboard entry is for \texttt{gpt-oss-120b}, \texttt{gpt-oss-20b} should be no higher.}%
    }
    \setcounter{footnote}{6}

    This mismatch matters beyond leaderboard rankings.
    When a user's coding tool wraps the same model in a different, unoptimized harness, the model severely underperforms its published numbers -- which may explain why some small coding LLMs that look strong on paper fail to gain traction in practice.

    \texttt{gpt-oss} compounds the problem further.
    Its paper uses a proprietary harness and discloses few details beyond a container tool. It also uses a custom chat format---Harmony.
    OpenAI's only published evaluation framework, \texttt{gpt-evals}, tests AIME25 \textit{without} tools, yet the paper reports far higher numbers \textit{with} tools and never specifies what those tools are.
    However, tools are crucial.
    On AIME25, \texttt{gpt-oss-20b} scores only 72.1\% without tools but 90.4\% with tools at MEDIUM reasoning -- tools make the difference between mediocre and near-perfect performance.
    And there are no other published results that reproduce OpenAI scores for \texttt{gpt-oss} with tools.\footnote{Despite it's name \cite{bi2025isgptoss} does not use tools in its evaluation.}
    To properly use tools, the model requires native Harmony chat format, which is significantly different from standard Chat Completions format.

    \section*{Tools}

    Why does this gap exist?
    The posted logs for \texttt{gpt-oss-120b} reveal a major failure mode: the model calls undefined tools.
    Specifically, it calls \texttt{apply\_patch} -- an undefined tool -- 860 times.\footnote{https://www.swebench.com/}
    This looks like hallucination, but \texttt{apply\_patch} is a built-in system tool\footnote{https://github.com/openai/gpt-oss} -- the behavior reflects a strong training prior, not a defect.
    Each time \texttt{gpt-oss} edits a file, it calls \texttt{apply\_patch} -- whether or not the tool is defined.
    So we added the \texttt{apply\_patch} tool, borrowed from Codex\footnote{https://github.com/openai/codex}, and this failure mode vanished.

    Are there any other in-distribution tools?
    We reverse-engineered the model's tool priors.
    The key insight: when prompted without any tool definitions, the model still attempts to call tools from its training distribution.
    We exploited this to discover tool names and schemas, then cross-referenced text descriptions with actual tool calls to filter confabulations from real parameters.

    \paragraph{Pipeline:}
    \begin{enumerate}
      \item We extracted 5683 intent--reasoning--tool tuples from raw agent logs.
      \item We fed 651 reasoning strings to the model with no tool definitions and recorded which tools it attempted to call; the top prior tools were \texttt{container\allowbreak.exec} (4.8\%), \texttt{repo\_browser\allowbreak.print\_tree} (3.4\%), and \texttt{repo\_browser\allowbreak.apply\_patch} (1.2\%).
      \item Targeted sampling with specific prompts pinned down argument schemas for each discovered tool (240 samples across four tool groups).
      \item We defined the discovered tools and validated them against no-tool baselines.
      \item We asked the model directly to list \texttt{repo\_browser.*} tools and cross-referenced text mentions with actual calls from all prior steps (160 samples).
      \item We forced the model to call each tool by name to collect canonical schemas with bootstrap confidence intervals (405 samples, 334 with tool calls).
      \item Finally, we asked the model to describe each tool's schema in natural language and filtered confabulated parameters against actual calls (405 samples).
    \end{enumerate}
    
    All confidence intervals reported below are 95\% bootstrap CIs (1,000 resamples).

    \paragraph{Cross-referencing text mentions with actual calls.}
    The model mentioned 170+ distinct \texttt{repo\_browser\allowbreak.*} tool names in text, most at under 2\% -- these are confabulations, not in-distribution tools.
    When we cross-reference text mention rates with actual tool calls from all baseline experiments, the real tools emerge (Table~\ref{tab:cross_ref}).

    \begin{table}[!htbp]
        \centering
        \begin{tabularx}{\linewidth}{LLLL}
          \hline
          \textbf{Tool} & \textbf{Text mentions [95\% CI]} & \textbf{Actual calls} & \textbf{Verdict} \\
          \hline
          \texttt{print\_tree} & 28.1\% [21.2, 35.0] & 101 & confirmed \\
          \texttt{search} & 34.4\% [27.5, 42.5] & 11 & confirmed \\
          \texttt{open\_file} & 44.4\% [36.9, 52.5] & 3 & confirmed \\
          \texttt{apply\_patch} & 2.5\% [0.6, 5.0] & 8 & confirmed \\
          \texttt{read\_file} & 25.0\% [18.8, 31.9] & 1 & likely alias \\
          \texttt{list\_files} & 10.6\% [6.2, 15.6] & 2 & likely alias \\
          \texttt{delete\_file} & 8.8\% [5.0, 13.1] & 0 & confabulated \\
          \texttt{write\_file} & 7.5\% [3.8, 11.9] & 0 & confabulated \\
          \hline
        \end{tabularx}
        \caption{Cross-reference of text-mentioned \texttt{repo\_browser} tools versus actual tool calls ($n=160$ text samples, $n=1201$ baseline samples). Tools with both text mentions and observed calls are confirmed; tools described but never called are confabulations.}
        \label{tab:cross_ref}
    \end{table}

    \paragraph{Schema robustness.}
    When explicitly asked to call each tool by name, 100\% of calls produce valid JSON.
    Table~\ref{tab:schema_robustness} shows the parameter presence rates for the three primary tools.
    Alias parameters such as \texttt{file\_path} (25\% in \texttt{open\_file}), \texttt{pattern} (5--7\% in \texttt{search}), and \texttt{start\_line}/\texttt{end\_line} (8\% in \texttt{read\_file}) appear at under 25\% and should be accepted but not required.
    The model confabulates 3--10 plausible parameters per tool (e.g., \texttt{encoding}, \texttt{case\_sensitive}, \texttt{recursive}) that it describes in text but never uses in actual calls.

    \begin{table}[!htbp]
        \centering
        \begin{tabularx}{\linewidth}{LLLL}
          \hline
          \textbf{Parameter} & \textbf{\texttt{print\_tree} ($n=108$)} & \textbf{\texttt{search} ($n=38$)} & \textbf{\texttt{open\_file} ($n=32$)} \\
          \hline
          \texttt{path} & 100\% [100, 100] & 97\% [92, 100] & 72\% [56, 88] \\
          \texttt{depth} & 99\% [97, 100] & --- & --- \\
          \texttt{query} & --- & 95\% [87, 100] & --- \\
          \texttt{max\_results} & --- & 53\% [37, 68] & --- \\
          \texttt{line\_start} & --- & --- & 78\% [62, 91] \\
          \texttt{line\_end} & --- & --- & 72\% [56, 88] \\
          \hline
        \end{tabularx}
        \caption{Schema robustness: parameter presence rates when the model is asked to call each tool by name (Step~5, 95\% bootstrap CIs).}
        \label{tab:schema_robustness}
    \end{table}

    \paragraph{Discovered tool inventory.}
    The model's in-distribution \texttt{repo\_browser} namespace contains three functional groups plus \texttt{apply\_patch} (Table~\ref{tab:tool_inventory}).
    Plus \texttt{container.exec} for shell execution.

    Alias tools such as \texttt{list\_dir}, \texttt{list\_directory}, \texttt{read\_file}, and \texttt{find} share schemas with their primary counterparts and collapse into the three core tools when tools are properly defined.

    \begin{table}[!htbp]
        \centering
        \begin{tabularx}{\linewidth}{LLLL}
          \hline
          \textbf{Group} & \textbf{Primary tool} & \textbf{Aliases} & \textbf{Canonical schema} \\
          \hline
          Directory listing & \texttt{print\_tree} & \texttt{list\_files}, \texttt{list\_dir} & \texttt{\{path, depth\}} \\
          Content search & \texttt{search} & \texttt{find} & \texttt{\{path, query, max\_results?\}} \\
          File reading & \texttt{open\_file} & \texttt{read\_file} & \texttt{\{path, line\_start?, line\_end?\}} \\
          Patch & \texttt{apply\_patch} & --- & \texttt{\{patch\}} \\
          \hline
        \end{tabularx}
        \caption{Discovered in-distribution \texttt{repo\_browser} tools with canonical schemas, plus \texttt{container.exec} as a flat tool.}
        \label{tab:tool_inventory}
    \end{table}

    \paragraph{Validation: baseline versus with tools.}
    Defining the discovered tools in the system message dramatically increases tool call rates (Table~\ref{tab:call_rates}).
    When we defined these tools in the developer message -- as OpenAI's cookbook documents\cite{kundel2025harmony} -- the model still failed to call them reliably.
    But when we moved them to the system message, the model called them reliably.

    \begin{table}[!htbp]
        \centering
        \begin{tabularx}{\linewidth}{LLLLL}
          \hline
          \textbf{Tool} & \textbf{Prompt type} & \textbf{Baseline} & \textbf{With tools} & \textbf{Lift} \\
          \hline
          \texttt{print\_tree} & explore & 29.4\% [20, 40] & 98.6\% [96, 100] & 3.4$\times$ \\
          \texttt{search} & search & 3.8\% [0, 9] & 58.8\% [48, 69] & 15$\times$ \\
          \texttt{open\_file} & read code & 1.2\% [0, 4] & 21.2\% [13, 30] & 18$\times$ \\
          \hline
        \end{tabularx}
        \caption{Tool call rates without tools defined (baseline) versus with tools in the system message (95\% bootstrap CIs).}
        \label{tab:call_rates}
    \end{table}

    \paragraph{Behavioral observations.}
    \texttt{print\_tree} dominates as the default first action even for search or file-reading prompts, matching the agent log pattern where directory listing is always the first command.
    \texttt{open\_file} has an overall call rate of only 21\% but reaches 60\% when the prompt contains both a file path and line numbers.
    \texttt{search} is reliably triggered by prompts with search-intent keywords such as ``search'', ``grep'', or ``find definition''.
    When the three primary tools are formally defined, alias variants (\texttt{list\_files}, \texttt{read\_file}, \texttt{find}) vanish -- the model uses the defined tools exclusively.

    \section*{Harmony agent}
  
    \begin{figure}[!htbp]
      \centering
      \includegraphics[width=0.8\textwidth]{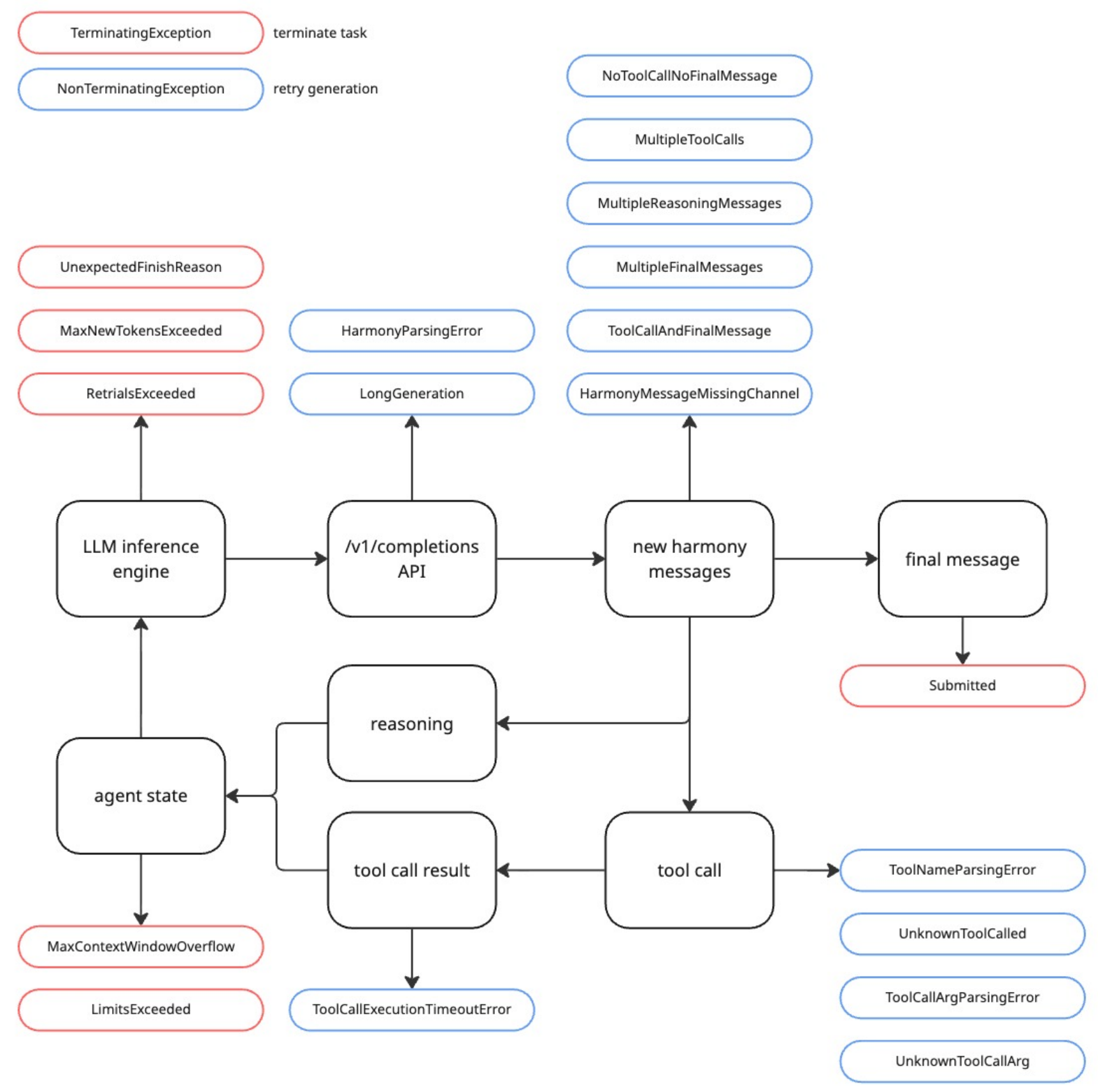}
      \caption{Harmony agent.}
      \label{fig:harmony_agent}
    \end{figure}

    A second major factor that explains the gap between published and reproduced scores: OpenAI trained \texttt{gpt-oss-20b} on the harmony format, which differs radically from the standard OpenAI Chat Completions format.
    Mapping one format to the other is complex.
    Yet no existing backend provides a native harmony API (to the best of our knowledge); they all convert to the OpenAI Chat Completions API or OpenAI Response API format, which may distort model output.
    Worse, none can define tools in harmony format.

    We solved this by implementing the Harmony Agent.
    We built on the simple yet powerful \texttt{mini\allowbreak-swe\allowbreak-agent}.\cite{yang2024sweagent}\footnote{https://github.com/SWE-agent/mini-swe-agent}

    Our harness encodes and decodes messages in harmony's native token format through the \texttt{openai\_harmony} library, so it never converts to Chat Completions format.
    The Harmony agent consumes the raw tokenized output of \texttt{gpt-oss}, i.e. \texttt{<|start|>{header}<|message|>{content}<|end|>}.
    It starts each run with three messages: a system message sets the model identity, reasoning effort, and tools in TypeScript-like syntax; a developer message defines the task instructions; a user message states the task.
    The agent then loops: render the conversation to tokens, query the model, parse the response, and either execute a tool call or terminate.

    Special tokens (\texttt{\textless|start|\textgreater}, \texttt{\textless|message|\textgreater}, \texttt{\textless|end|\textgreater}) delimit each message; every message carries a role and a channel.
    The model responds on three channels: \texttt{analysis} for chain-of-thought reasoning, \texttt{commentary} for tool calls, and \texttt{final} for the user-facing answer.
    \texttt{<final>} message terminates the task.
    Because channels separate message types, the harness routes them accordingly.

    \begin{figure}[!htbp]
      \centering
      \includegraphics[width=\textwidth]{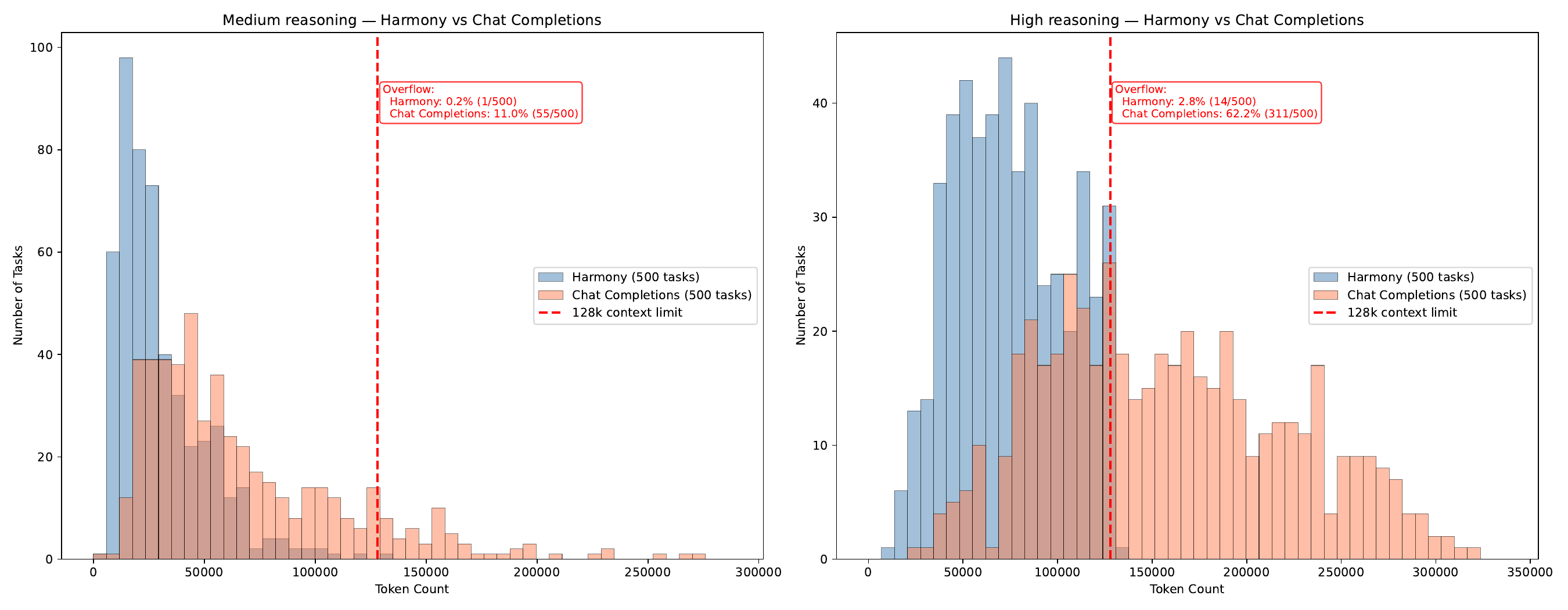}
      \caption{Harmony vs chat completions token usage.}
      \label{fig:token_histograms}
    \end{figure}

    In harmony, the system or developer message defines tools once, saving many tokens; the OpenAI Chat Completions format repeats them every turn.
    Figure~\ref{fig:token_histograms} shows token usage across all Harmony trajectories for 500 SWE Verified instances with \texttt{gpt-oss-20b} at Medium and High reasoning.
    For Chat Completions estimates, we simply add the number of tokens for tool definitions to each turn.
    This difference is especially pronounced for High reasoning.
    
    When the model calls a tool, the harness reads the tool name from the recipient field, parses the JSON arguments, executes the tool in a sandboxed container, and appends the result as a tool-role message.
    The harness validates each step and raises corresponding custom exceptions.
    Exceptions fall into two categories (following the SWE-bench paper): Terminating and NonTerminating.
    Terminating exceptions halt the run because the harness cannot recover by retrying.
    The harness retries malformed calls and unparseable arguments up to ten times, but terminates the run when the context window overflows or the agent exceeds its step limit.
    NonTerminating exceptions signal errors the model can likely recover from on retry.

    NonTerminatingException:
    \begin{itemize}
      \item \texttt{LongGeneration}: Raised when generation exceeded \texttt{max\_tokens}.
      \item \texttt{HarmonyParsingError}: Harmony error when parsing LLM completion.
      \item \texttt{HarmonyMessageMissingChannel}: Raised when a non-system harmony message is missing a channel.
      \item \texttt{MultipleReasoningMessages}: Raised when multiple reasoning messages.
      \item \texttt{MultipleFinalMessages}: Raised when multiple final messages are found.
      \item \texttt{MultipleToolCalls}: Multiple tool calls.
      \item \texttt{NoToolCallNoFinalMessage}: Raised when no tool call and no final message are received.
      \item \texttt{ToolCallAndFinalMessage}: Raised when both tool call and final message are received.
      \item \texttt{ToolNameParsingError}: Raised when tool name failed to parse.
      \item \texttt{UnknownToolCalled}: Unknown tool called.
      \item \texttt{UnknownToolCallArg}: Tool call has unknown argument.
      \item \texttt{ToolCallArgParsingError}: Failed to parse tool call arguments.
      \item \texttt{ExecutionTimeoutError}: Raised when the action execution timed out.
    \end{itemize}
    TerminatingException:

    \begin{itemize}
      \item \texttt{Submitted}: Raised when the LM declares that the agent has finished its task.
      \item \texttt{LimitsExceeded}: Raised when the agent has reached its cost or step limit.
      \item \texttt{MaxContextWindowOverflow}: Raised when max context window of the model is reached.
      \item \texttt{UnexpectedFinishReason}: Raised when OpenAI completions finishes with unknown reason.
      \item \texttt{MaxNewTokensExceeded}: Raised when \texttt{max\_tokens} is exceeded.
      \item \texttt{RetrialsExceeded}: Raised when retrials on \texttt{NonTerminatingException}s are exceeded.
    \end{itemize}

    This exception system is crucial: OpenAI recommends running \texttt{gpt-oss} at \texttt{temperature} 1 and \texttt{top\_p} 1\footnote{https://github.com/openai/gpt-oss}, so the model may deviate from the harmony format.
    If we apply the harmony format and use in-distribution tools, the model rarely deviates, and \texttt{gpt-oss} recovers on retry.

    \section*{Evaluation}
    
    The real test of our approach is how well we can reproduce the published results.
    We ran two benchmarks -- AIME25 and SWE Verified -- and achieved results that match those in OpenAI's paper on \texttt{gpt-oss-20b}.
    All experiments were run with \texttt{vLLM 0.14.1} inference with: \texttt{--tensor-parallel-size 1 --max-model-len 131072}.
    All results are pass@1.
    In the HIGH reasoning setting for SWE Verified, we only retried \texttt{MaxContextWindowOverflow} exceptions until the agent reached a final message, since verbose reasoning traces often overflow the 128k context window.
    We could alternatively compact the context, but that approach is more complex and we left it for future work.

    \begin{table}[!htbp]
        \centering
        \begin{tabularx}{\linewidth}{LLLL}
          \hline
          \textbf{Benchmark} & \textbf{Published score\cite{gptoss}} & \textbf{HarmonyAgent score} & \textbf{95\% Bootstrap CI} \\
          \hline
          SWE Verified HIGH reasoning & 60.7\% & 60.4\% & [56.2\%, 64.8\%] \\
          SWE Verified MEDIUM reasoning & 53.2\% & 53.3\% & [49.3\%, 57.7\%] \\
          AIME 2025 MEDIUM reasoning with tools & 90.4\% & 91.7\% & [87.5\%, 95.0\%] \\
          \hline
        \end{tabularx}
        \caption{OpenAI published results vs HarmonyAgent results (with 95\% bootstrap CIs, 1,000 resamples).}
        \label{tab:eval_results}
    \end{table}

    \section*{Conclusions}

    The gap between published and reproduced scores for \texttt{gpt-oss-20b} is not a model deficiency -- it is a harness deficiency.
    We closed that gap with two changes: we provided the model's in-distribution tools (\texttt{container.exec}, \texttt{repo\_browser} and \texttt{apply\_patch}), and we implemented a native Harmony harness rather than relying on Chat Completions format.
    Our scores match OpenAI's published numbers within confidence intervals on both SWE Verified and AIME25.

    This result generalizes.
    Any model trained on a specific tool set and message format might underperform when a generic harness omits those tools or converts that format.
    This is especially true for \texttt{gpt-oss}: the 36-point gap between \texttt{gpt-oss-120b}'s published score and its SWE leaderboard entry is the expected cost of harness mismatch.
    When practitioners deploy open-source models in coding agents and see disappointing results, the model may not be at fault -- the harness may be.

    Without an open harness, practitioners dismissed \texttt{gpt-oss-20b} as weaker than it is.
    Open-source models deserve correct open-source evaluation harnesses.

    \bibliographystyle{plainnat} % Choose a bibliography style
    \bibliography{references} % Specify your BibTeX file (references.bib)
\end{document}